\documentclass{article}

\pdfoutput = 1

\PassOptionsToPackage{numbers, compress}{natbib}

\usepackage[nonatbib,preprint]{neurips_2019}

\usepackage[utf8]{inputenc} 
\usepackage[T1]{fontenc}    
\usepackage{adjustbox}
\usepackage{booktabs}
\usepackage{multirow}
\usepackage{amsmath}
\usepackage{mathtools}
\usepackage{float}
\usepackage{listings}
\usepackage{csquotes}
\usepackage[dvipsnames]{xcolor}
\usepackage{subfigure}
\usepackage{url}            
\usepackage{booktabs}       
\usepackage{amsfonts}       
\usepackage{nicefrac}       
\usepackage{microtype}      

\definecolor{codegreen}{rgb}{0,0.6,0}
\definecolor{codegray}{rgb}{0.5,0.5,0.5}
\definecolor{codepurple}{rgb}{0.58,0,0.82}
\definecolor{backcolour}{rgb}{0.95,0.95,0.92}
 
\lstdefinestyle{mystyle}{
    backgroundcolor=\color{backcolour},   
    commentstyle=\color{codegreen},
    keywordstyle=\color{magenta},
    numberstyle=\tiny\color{codegray},
    stringstyle=\color{codepurple},
    basicstyle=\ttfamily\footnotesize,
    breakatwhitespace=false,         
    breaklines=true,                 
    captionpos=b,                    
    keepspaces=true,                 
    numbers=left,                    
    numbersep=5pt,                  
    showspaces=false,                
    showstringspaces=false,
    showtabs=false,                  
    tabsize=2
}

\newfloat{lstfloat}{htbp}{lop}

\DeclarePairedDelimiter\floor{\lfloor}{\rfloor}
\DeclarePairedDelimiter\abs{\lvert}{\rvert}%

\title{Revisiting Classical Bagging with Modern Transfer Learning for On-the-fly Disaster Damage Detector}

\author{%
  Junghoon Seo\thanks{Both authors contributed equally to this manuscript.}\\
  Satreci Initiative Co., Ltd.\\
  Daejeon, Rep. of Korea\\
  \texttt{sjh@satreci.com}\\
  \And
  Seungwon Lee\footnotemark[1]\hspace{1.5mm}\thanks{This work was done while Seungwon Lee and Beomsu Kim were interns at SI Analytics Co., Ltd..}\\
  School of Materials Science and Engineering, GIST\\
  Gwangju, Rep. of Korea\\
  \texttt{lsdras96@gist.ac.kr} \\
  \AND
  Beomsu Kim\footnotemark[2]\\
  Department of Mathematical Sciences, KAIST\\
  Daejeon, Rep. of Korea\\
  \texttt{3141kbs@kaist.ac.kr} \\
  \And
  Taegyun Jeon\\
  SI Analytics Co., Ltd.\\
  Daejeon, Rep. of Korea\\
  \texttt{tgjeon@si-analytics.ai}
}

\begin{document}

\maketitle

\begin{abstract}
Automatic post-disaster damage detection using aerial imagery is crucial for quick assessment of damage caused by disaster and development of a recovery plan.
The main problem preventing us from creating an applicable model in practice is that damaged (positive) examples we are trying to detect are much harder to obtain than undamaged (negative) examples, especially in short time.
In this paper, we revisit the classical bootstrap aggregating approach in the context of modern transfer learning for data-efficient disaster damage detection.
Unlike previous classical ensemble learning articles, our work points out the effectiveness of simple bagging in deep transfer learning that has been underestimated in the context of imbalanced classification.
Benchmark results on the AIST Building Change Detection dataset show that our approach significantly outperforms existing methodologies, including the recently proposed disentanglement learning.
\end{abstract}

\section{Introduction}
Automatic post-disaster damage detection is one of the most important topics in remote sensing.
It is crucial to quickly monitor damages caused by various types of disaster and develop a recovery plan.
Approaches propsed by many previous studies require a large-scale satellite image dataset.
For example, \cite{gueguen2015large} collected a multi-source satellite dataset to train a binary classifier with a human-in-the-loop scheme.
This dataset covers a total area of 4665 km$^2$, and the area of significant damages labeled by expert photo-interpreters of UNITAR/UNOSAT is over 174 km$^2$.
\cite{fujita2017damage} obtained several pairs of pre- and post-tsunami aerial images covering 66 km$^2$ of tsunami-affected areas from the PASCO image archive and conducted a survey assessing over 220,000 buildings in the ravaged areas.
Another work \cite{doshi2018satellite} also required large-scale datasets like Spacenet dataset \cite{van2018spacenet} and DeepGlobe dataset \cite{demir2018deepglobe} to train a classifier for detecting man-made features in satellite images.

However, the requirement for large-scale datasets is impractical in many real world problems for two major reasons.
First, in most cases, it is crucial to receive the results of the damage detection model immediately after the disaster.
Decision makers responsible for taking action on the disaster often do not have enough time to wait for collection of datasets.
Second, before the disaster, we may not be able to specify objects for damage assessment or  causes of the damage.
It is quite difficult to build datasets that are compatible with all kinds of objects and causes, and that can be generalized to any satellite sensor.

Therefore, in this work, we propose an approach to train a model on-the-fly on a minimally labeled dataset which is constructed after the disaster.
In this case, we need to address \emph{class rarity} and \emph{class imbalance}.
From a chronological sequence of satellite images, we can easily get numerous undamaged (negative) pairs.
In contrast, the number of the damaged (positive) examples is extremely small because damaged examples are collected in a short time after the disaster.
To alleviate class rarity and class imbalance, we formulate post-disaster damage detection as an extremely imbalanced classification problem.

We revisit classical bootstrap aggregating (bagging) approach with modern transfer learning to solve class rarity and class imbalance.
Our methodology is motivated by insights of previous researches on ensemble learning or transfer learning.
Note that existing methods compared in this paper focused on properly pre-training models using a large number of negative pairs or making under-sampled datasets to balance positive and negative data.
We instead emphasize the importance of making full use of imbalanced binary labeled datasets in the fine-tuning process.
Despite the methodological simplicity, our method shows good performance in the AIST Building Change Detection dataset. Concretely, our contributions in this paper are the following:
\begin{enumerate}
  \item We revisit the bagging-based ensemble method in deep transfer learning in the context of extremely imbalanced classification, whose effectiveness is underestimated in the era of deep learning. We also discuss the rationale behind the power of our approach.
  \item In an extremely imbalanced classification setting, our simple ResNet34 ensemble model outperforms previous state-of-the-art models in the AIST Building Change Detection dataset by a significantly large margin. We also discuss the effect of ensemble size on transfer learning performance.
\end{enumerate}

\section{Motivation}
\label{sec:mot}
Our methodology has a 3-fold motivation: the representation power of ImageNet pre-trained models, maximization of data usage with ensemble method, and auto-calibration property of transfer learning models.
The empirical justification for this motivation is discussed in Section \ref{sec:abs}.

First, it is well known that transfer learning from an ImageNet-pretrained model stabilizes training process and generally yields higher performance.
There are some works \cite{sun2017revisiting, he2019rethinking, raghu2019transfusion} which indicated that using a pre-training model has no performance benefit when sufficient data are available.
Nevertheless, transfer learning from ImageNet-pretrained models is still a popular hard baseline of many tasks, especially in a few-shot setting \cite{chen2019closer, chen2018lstd}.
The baseline of transfer learning was also presented in the benchmark of our target dataset \cite{hamaguchi2019rare}, and showed results competitive to those of a sophisticated disentanglement learning framework presented in \cite{hamaguchi2019rare}.

Second, simple under-sampling scheme used in  existing works about post-disaster damage detection \cite{gueguen2015large, hamaguchi2019rare} did not give the model sufficient supervision during the fine-tuning phase.
This causes low accuracy and high variance in terms of performance of damage detection because model convergence in the fine-tuning phase depends on how negative data are sampled.
In the context of classical machine learning framework, it is well known that methods on bagging-based imbalanced classification  effectively alleviate the two same problems, low accuracy and high variance of model performance.
Thus, we adopted these classical bagging approaches to construct simple ensemble baselines which make maximum usage of negative data.

Lastly, it has recently been found that models transferred from an ImageNet-pretrained model have self-calibration properties \cite{hendrycks2019using} that are advantageous in ensemble learning.
This is important in our scheme because the effect of ensembling depends on how well each base model used in the ensemble is calibrated \cite{bella2013effect, lakshminarayanan2017simple}.
To demonstrate the presence of the calibration effect in transfer learning, in Section \ref{sec:abs}, we measured calibration error metric values \cite{hendrycks2019deep} of a total of 800 base models (200 base models in each experiment setting).
Results of the measurement are shown in Table \ref{tab:cal}.
At positive 5-shot settings, using an ImageNet-pretrained model achieved significantly lower calibration error than a scratch model.
At positive 50-shot settings, however, this difference became insignificant.
Based on observations of existing studies and results from these measurements, we anticipated that the effectiveness of the model ensemble would depend on the degree of calibration of the base model.

\begin{table}[]
\begin{tabular}{c|cc|cc}
\multirow{2}{*}{} & \multicolumn{2}{c|}{\# Labels 5} & \multicolumn{2}{c}{\# Labels 50} \\ \cline{2-5} 
 & \multicolumn{1}{l}{RMS Error (\%)} & \multicolumn{1}{l|}{MAD Error (\%)} & \multicolumn{1}{l}{RMS Error (\%)} & \multicolumn{1}{l}{MAD Error (\%)} \\ \hline
\begin{tabular}[c]{@{}c@{}}LeNet\\ (Scratch)\end{tabular} & \begin{tabular}[c]{@{}c@{}}22.06\\ ($\pm$6.57)\end{tabular} & \begin{tabular}[c]{@{}c@{}}18.71\\ ($\pm$6.00)\end{tabular} & \begin{tabular}[c]{@{}c@{}}7.98\\ ($\pm$2.45)\end{tabular} & \begin{tabular}[c]{@{}c@{}}6.18\\ ($\pm$1.96)\end{tabular} \\
\begin{tabular}[c]{@{}c@{}}ResNet34\\ (Pre-trained)\end{tabular} & \begin{tabular}[c]{@{}c@{}}12.79\\ ($\pm$3.50)\end{tabular} & \begin{tabular}[c]{@{}c@{}}10.52\\ ($\pm$2.87)\end{tabular} & \begin{tabular}[c]{@{}c@{}}7.60\\ ($\pm$1.34)\end{tabular} & \begin{tabular}[c]{@{}c@{}}6.53\\ ($\pm$1.09)\end{tabular}
\end{tabular}
\caption{Calibration errors for models trained from scratch (LeNet) and models with pre-training (ResNet34). Detailed descriptions for each model can be found in Section \ref{sec:pro}.}
\label{tab:cal}
\end{table}

\section{Experiments}
\subsection{Dataset and Benchmark Methods}
We used the AIST Building Change Detection (ABCD) dataset \cite{fujita2017damage} as the benchmark dataset.
ABCD dataset covers the wake of the Great East Japan earthquake on March 11, 2011 and each  pair of the dataset is labeled as 'building damaged' (positive) or 'building not damaged' (negative).
\emph{Resized} version of the ABCD dataset is comprised of 6,848 negative data and 4,546 positive data with $120 \times 120$ image size.
For more details about the dataset, please refer to the original dataset paper \cite{fujita2017damage}.
Positive and negative samples from the ABCD dataset are shown in Appendix \ref{app:abcd}.

For performance comparison among multiple models on the ABCD dataset, we gathered a list of methods from previous benchmark results: simple under/over-sampling scheme, transfer learning from an ImageNet-pretrained VGG model \cite{simonyan2015very}, Multi-Level Variational Auto-Encoder (MLVAE) \cite{bouchacourt2018multi}, Mathieu et al. \cite{mathieu2016disentangling}, and Hamaguchi et al. \cite{hamaguchi2019rare}.
It is noteworthy that the latter three methods concentrated on representation learning using only negative pairs based on sophisticated disentanglement learning.
We collected benchmark results of these methods from \cite{hamaguchi2019rare}.

To simulate class rarity and class imbalance of positive data, a small number of positive data sampled without replacement were used in the training phase.
For example, the \enquote{\emph{positive 50-shot}} setting means that only 50 samples out of a total of 4,546 positive samples will be used in training.
In contrast, all 6,848 negative samples can always be used regardless of experiment settings.

\subsection{Details of Experiment Procedure and Setting}
\label{sec:pro}
To construct simple ensemble baselines, we followed the partitioning approach of \cite{chan1998toward, yan2003predicting}.
Like other bagging-based imbalanced classification methods, partitioning method was proposed to compensate for under-sampling or over-sampling in the class imbalance situation.
The following procedure was used for model training and inference:
\begin{enumerate}
  \item Sample positive dataset $P$ without replacement. The sampled positive dataset is notated as $P^\prime$. Then, shuffle negative dataset $N$ and split negative data into chunks whose size equals the number of positive samples. Each chunk is indexed by a number starting at 1 and ending at the number of chunks, i.e., $N = N_1 \cup N_2 \cup \cdots \cup N_{\floor*{\abs{N} / \abs{P^\prime}}}$. For example, in the positive 50-shot setting, the negative dataset is divided into $\floor*{\frac{6,848}{50}} = 136$ chunks.
  \item Sample negative chunks by the size of model ensemble $\abs{M}$ without replacement. Each sampled negative chunk $N^\prime$ is indexed by a number starting at 1 and ending at $\abs{M}$.
  \item Train each base classification model $M_i$ using $D_i =  P^\prime \cup N^\prime_i$. For inference, feed the sample to each $M_i$ and get $\abs{M}$ scores from the base models. Then, use the average of all the $\abs{M}$ scores as the final decision score.
\end{enumerate}

We used the ResNet34 backbone \cite{he2016deep} as the baseline of the ImageNet-pretrained model which has the siamese architecture in \cite{hamaguchi2019rare}.
For the base model without transfer learning, we observed that the ResNet34 model  tended to overfit excessively on the training data, so a modified LeNet \cite{lecun1998gradient} was used instead.\footnote{The implementation code of this modified LeNet in PyTorch framework can be found in Appendix \ref{app:arch}.}
Feature size of the penultimate layer for both ResNet34 and LeNet architecture was fixed at 128.
Using pre-trained weights, the model converged within 20 iterations for positive 5-shots and 50 iterations for positive 50-shots.
From scratch, the model converged within 100 iterations for positive 5-shots and 130 iterations for positive 50-shots.
Considering the stochastic effect in the training process, each result of both base models ($\abs{M} = 1$) and the ensemble models ($\abs{M} > 1$) was averaged over 200 trials.
Adam optimizer \cite{kingma2015adam} was used with learning rate of 0.001 for training, and label smoothing technique was applied to facilitate model calibration \cite{muller2019does} with $\alpha = 0.1$ and $T = 0.0$.

\subsection{ABCD dataset Benchmark Results}

\begin{table}[t]
\begin{adjustbox}{center}
\begin{tabular}{c|cccccc|c}
\multicolumn{1}{l|}{} & \multicolumn{6}{c|}{Results from \cite{hamaguchi2019rare}} & \multicolumn{1}{c}{Results of this work} \\ \cline{2-8}
 & \begin{tabular}[c]{@{}c@{}}Under\\ -sampling\end{tabular} & \begin{tabular}[c]{@{}c@{}}Over\\ -sampling\end{tabular} & \begin{tabular}[c]{@{}c@{}}Transfer\\ (VGG)\end{tabular} & \begin{tabular}[c]{@{}c@{}}MLVAE\\ \cite{bouchacourt2018multi}\end{tabular} & \begin{tabular}[c]{@{}c@{}}Mathieu et al.\\ \cite{mathieu2016disentangling}\end{tabular} & \begin{tabular}[c]{@{}c@{}}Hamaguchi et al.\\ \cite{hamaguchi2019rare}\end{tabular} & Ours \\ \hline
\#Labels 5 & \begin{tabular}[c]{@{}c@{}}61.14\\ ($\pm$11.61)\end{tabular} & \begin{tabular}[c]{@{}c@{}}60.88\\ ($\pm$13.58)\end{tabular} & \begin{tabular}[c]{@{}c@{}}77.39\\ ($\pm$7.30)\end{tabular} & \begin{tabular}[c]{@{}c@{}}65.36\\ ($\pm$5.19)\end{tabular} & \begin{tabular}[c]{@{}c@{}}64.73\\ ($\pm$5.41)\end{tabular} & \begin{tabular}[c]{@{}c@{}}78.52\\ ($\pm$5.01)\end{tabular} & 
\begin{tabular}[c]{@{}c@{}}89.96\\ ($\pm$0.84)\end{tabular} \\
\#Labels 50 & \begin{tabular}[c]{@{}c@{}}64.05\\ ($\pm$17.16)\end{tabular} & \begin{tabular}[c]{@{}c@{}}54.05\\ ($\pm$11.78)\end{tabular} & \begin{tabular}[c]{@{}c@{}}88.17\\ ($\pm$0.75)\end{tabular} & \begin{tabular}[c]{@{}c@{}}86.31\\ ($\pm$1.80)\end{tabular} & \begin{tabular}[c]{@{}c@{}}77.66\\ ($\pm$2.11)\end{tabular} & \begin{tabular}[c]{@{}c@{}}89.70\\ ($\pm$0.77)\end{tabular} &
\begin{tabular}[c]{@{}c@{}}92.56\\ ($\pm$0.62)\end{tabular} \\ \hline
\end{tabular}
\end{adjustbox}
\caption{Benchmark results on the ABCD dataset. We report the mean and corresponding standard deviation of accuracies. \emph{\#Labels 5} and \emph{\#Labels 50} mean that 5 and 50 positive data were used, respectively.}
\label{tab:bench}
\end{table}

Table \ref{tab:bench} shows ABCD dataset benchmark results of the existing models and some of our models.
For our work, we report the results of the ensemble model using 20 base models with transfer training for both positive 5-shot and 50-shot case.
Our methodology, despite its simplicity, outperformed all previous benchmark models in terms of accuracy.
Specifically, our methodology improved average error rate by $ 1- \frac{10.04}{21.48} = 53.3\% $ for the positive 5-shot case and $1 - \frac{7.44}{10.3} = 27.8\% $ for the positive 50-shot case, compared to the existing state-of-the-art model.
Since this gap is larger in the positive 5-shot than in the positive 50-shot, our partitioning scheme is particularly effective when there are fewer positive samples.
It is also noteworthy that our models not only have high average accuracies, but also have smaller performance variances.
Since we can reliably obtain high accuracies in positive few-shot situations, our methodology is suitable for creating on-the-fly post-disaster damage detectors in real world applications.
\subsection{Ablation Studies}
\label{sec:abs}
\begin{figure}[t!p]
\begin{adjustbox}{center}
\subfigure[]{\includegraphics[width=.5\linewidth]{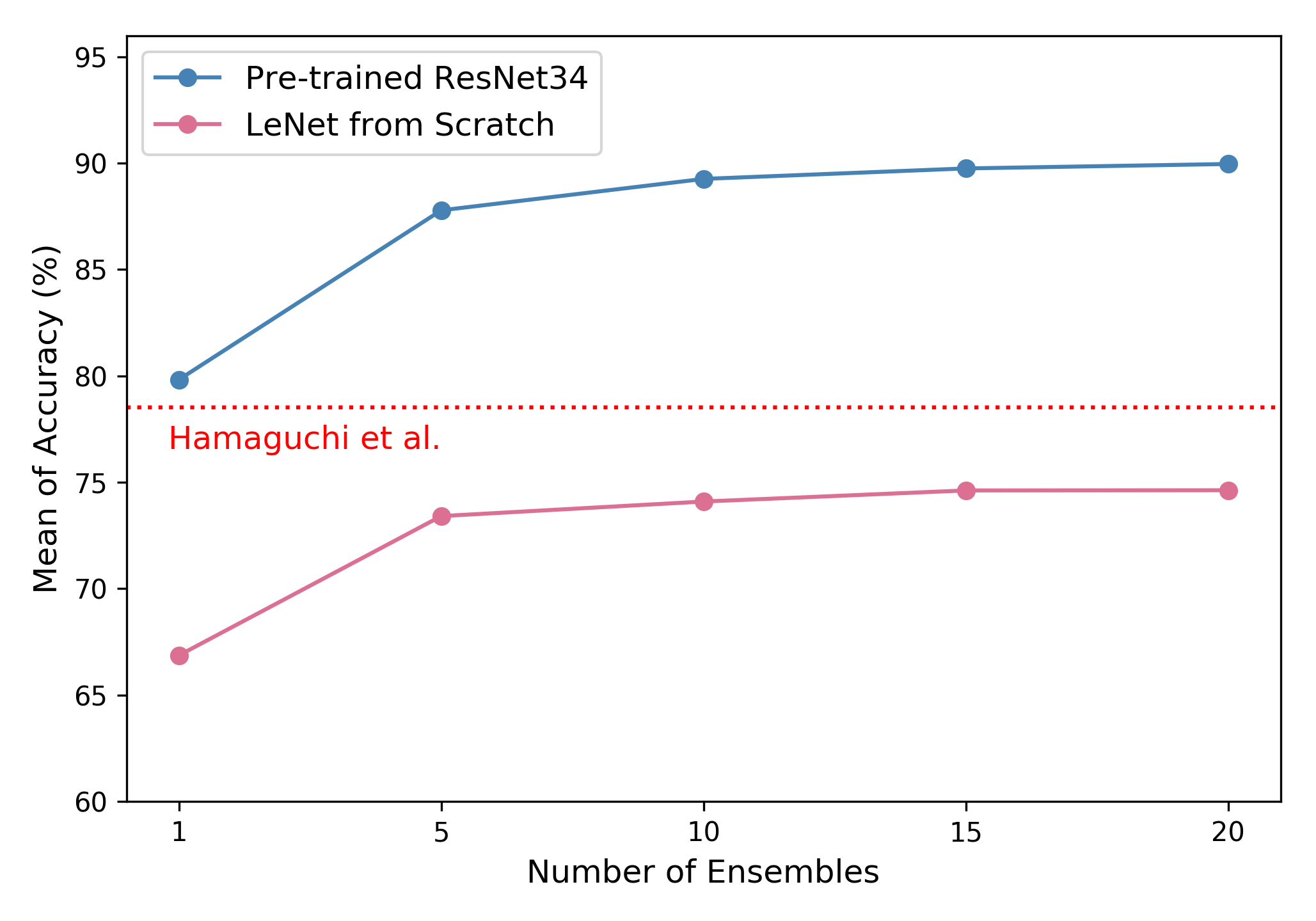}}
\subfigure[]{\includegraphics[width=.5\linewidth]{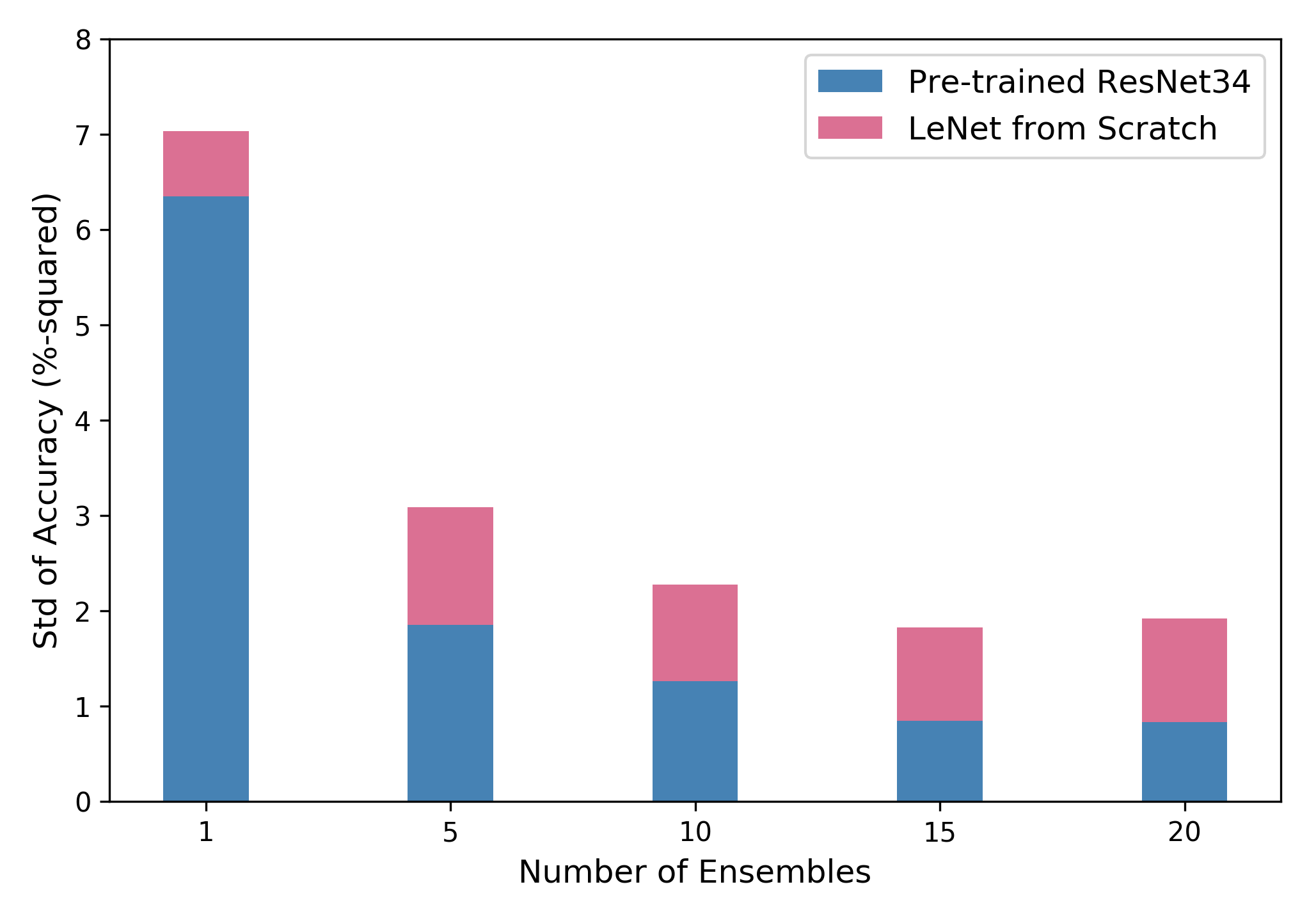}}
\end{adjustbox}
\caption{Mean ((a), Left) and standard deviation ((b), Right) of accuracies on the ABCD dataset with 5 positive data.}
\label{fig:5}
\end{figure}

\begin{figure}[t!p]
\begin{adjustbox}{center}
\subfigure[]{\includegraphics[width=.5\linewidth]{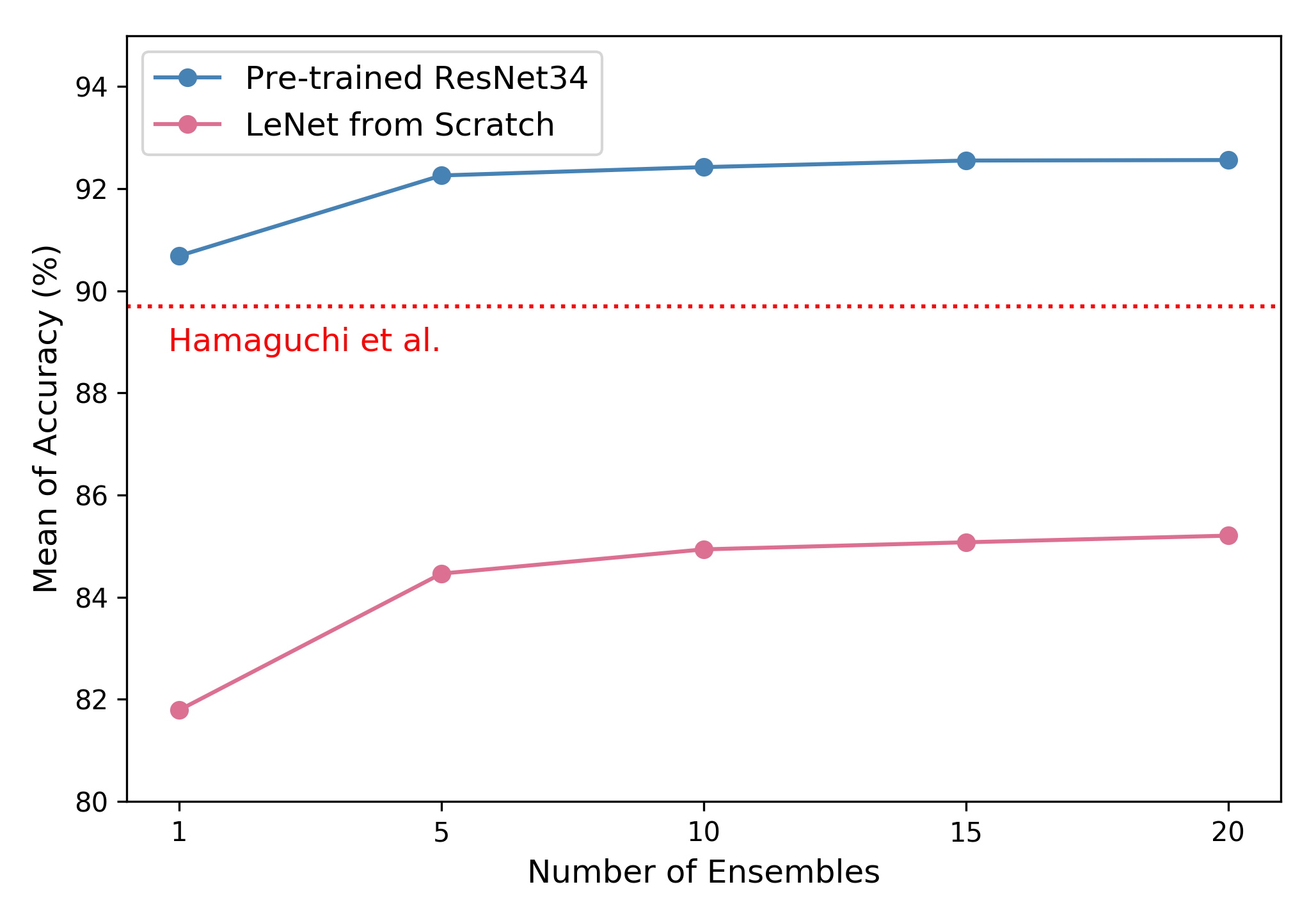}}
\subfigure[]{\includegraphics[width=.5\linewidth]{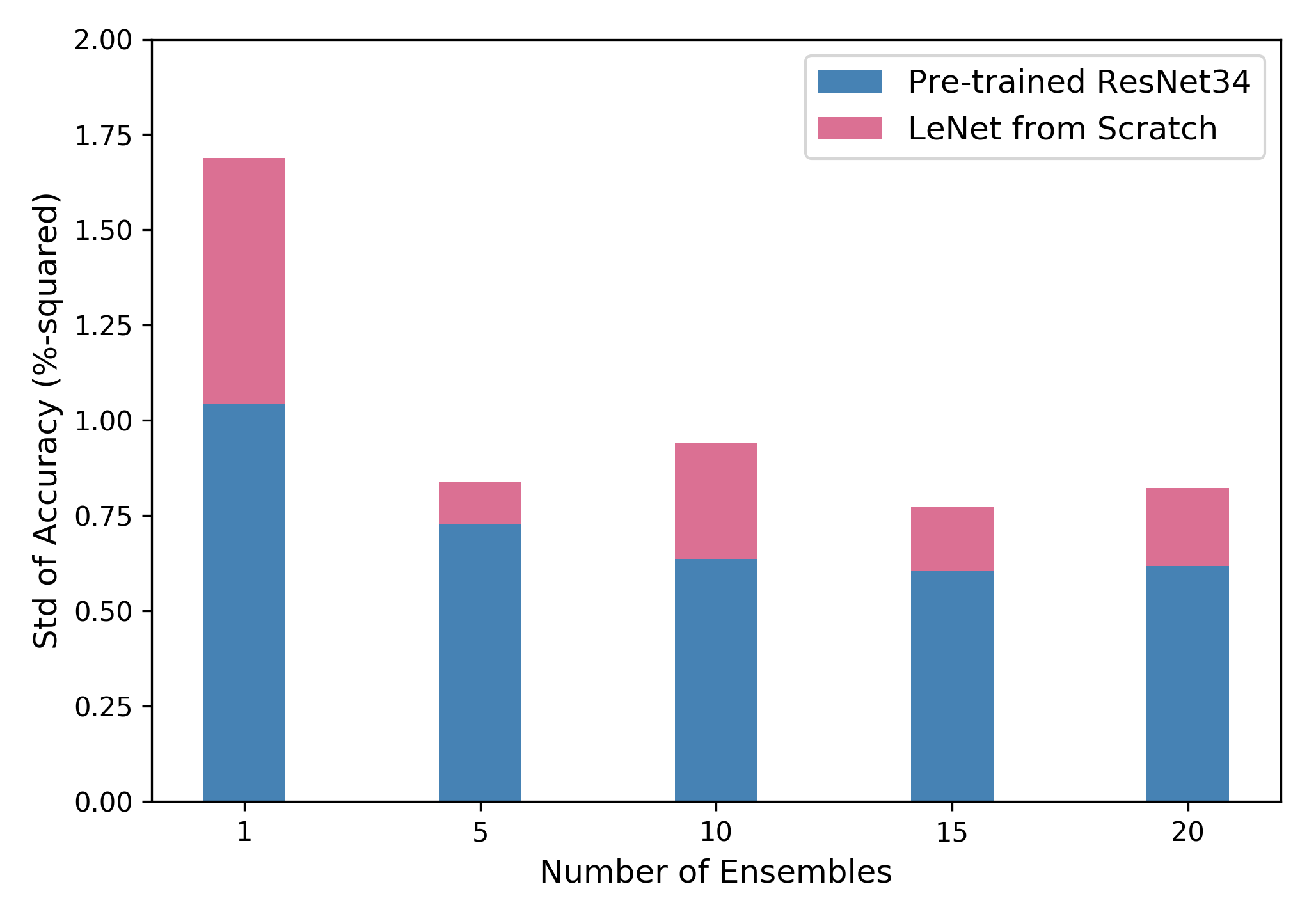}}
\end{adjustbox}
\caption{Mean ((a), Left) and standard deviation ((b), Right) of accuracies on the ABCD dataset with 50 positive data.}
\label{fig:50}
\end{figure}

To examine the effects of transfer learning and model ensembling, we conducted a 2-fold ablation study.
First, we trained ensemble models with base model as the pre-trained Resnet34 model (transfer learning) or the modified LeNet model (training from scratch).
Second, we trained ensemble models using various ensemble size.
The number of base models for ensemble models is set to 1, 5, 10, 15, and 20.
Fig. \ref{fig:5} and Fig. \ref{fig:50} show results from positive 5-shot and 50-shot setting for the ablation study.

For both positive 5-shot and 50-shot cases, base models transferred from ImageNet-pretrained weights ensured higher performance and lower variance than base models trained from scratch.
This result is quite natural in the light of existing discussions on the effectiveness of pre-training in low-shot settings.

Interestingly, in terms of error rate, the benefits of transfer learning is larger in a positive 5-shot setting than in a positive 50-shot setting.
For instance, comparing the error rate from single base model to 5-size ensemble model in the positive 5-shot setting (Fig. \ref{fig:5} (a), $\abs{M} = 1$ and $\abs{M} = 5$), the improvement was $1 - \frac{12.2}{20.2} = 39.6\%$ for the pre-trained base model and $1 - \frac{26.6}{33.1} = 19.6\%$ for the scratch base model.
On the other hand, in the case of the positive 50-shot case (Fig. \ref{fig:50} (a), $\abs{M} = 1$ and $\abs{M} = 5$), the improvement was $1 - \frac{7.7}{9.3} = 17.2\%$ for the pre-trained base model and $1 - \frac{15.5}{18.2} = 14.8\%$ for the scratch base model.
This observation is also valid for the variance of accuracy.
In the positive 5-shot ensemble setting (Fig. \ref{fig:5} (b), $\abs{M} > 1$), using a pre-trained model appears to reduce the variance up to about half of the variance of the scratch model.
However, this variance reduction effect is significantly reduced at the 50-shot setting (Fig. \ref{fig:50} (b), $\abs{M} > 1$).
This observation is well aligned with the motivation about model calibration, which is discussed in Section \ref{sec:mot} and Table \ref{tab:cal}.

As for the number of base models in the ensemble, we obtain trivial conclusions:\footnote{One thing to notice is that single models in our setting has much higher performance than those presented in previous benchmarks. This means that the effectiveness of the under-sampling scheme was under-estimated in existing benchmarks.
This is probably because the ResNet34 and LeNet architecture used in this work are more suitable for each experimental setting than the previously used VGG architecture.} as the number of base models used for ensembling increases, the ensemble model tends to achieve higher accuracy and lower variance.
In addition, when using more than 5 base models, the improvements of model performance and variance are rapidly saturated with respect to the number of base models.
Therefore, the use should consider the loss in memory or processing time and gain in performance-variance from model ensembling when selecting the number of base models.

\section{Conclusions}
In this paper, we pointed out problems with existing post-disaster damage detection frameworks and re-formulated the damage detection task as an imbalanced few-shot classification problem.
Instead of creating a large-scale dataset for damage detection, we proposed a methodology to effectively solve the imbalanced few-shot classification problem by combining classical bagging and modern transfer learning.
Our methodology complemented the blind spots of existing under-sampling schemes and showed superior performance compared to existing state-of-the-art models on the ABCD dataset.
We expect this methodology to potentially become a new reference approach for automatic post-disaster damage detection.


\newpage

\small

\bibliographystyle{unsrt}
\bibliography{neurips_2019.bbl}

\begin{thebibliography}{10}

\bibitem{gueguen2015large}
Lionel Gueguen and Raffay Hamid.
\newblock Large-scale damage detection using satellite imagery.
\newblock In {\em CVPR}, pages 1321--1328, 2015.

\bibitem{fujita2017damage}
Aito Fujita, Ken Sakurada, Tomoyuki Imaizumi, Riho Ito, Shuhei Hikosaka, and
  Ryosuke Nakamura.
\newblock Damage detection from aerial images via convolutional neural
  networks.
\newblock In {\em MVA}, 2017.

\bibitem{doshi2018satellite}
Jigar Doshi, Saikat Basu, and Guan Pang.
\newblock From satellite imagery to disaster insights.
\newblock In {\em NIPS Workshop}, 2018.

\bibitem{van2018spacenet}
Adam Van~Etten, Dave Lindenbaum, and Todd~M Bacastow.
\newblock Spacenet: A remote sensing dataset and challenge series.
\newblock {\em arXiv preprint arXiv:1807.01232}, 2018.

\bibitem{demir2018deepglobe}
Ilke Demir, Krzysztof Koperski, David Lindenbaum, Guan Pang, Jing Huang, Saikat
  Basu, Forest Hughes, Devis Tuia, and Ramesh Raska.
\newblock Deepglobe 2018: A challenge to parse the earth through satellite
  images.
\newblock In {\em CVPR Workshop}, 2018.

\bibitem{sun2017revisiting}
Chen Sun, Abhinav Shrivastava, Saurabh Singh, and Abhinav Gupta.
\newblock Revisiting unreasonable effectiveness of data in deep learning era.
\newblock In {\em ICCV}, pages 843--852, 2017.

\bibitem{he2019rethinking}
Kaiming He, Ross Girshick, and Piotr Doll{\'a}r.
\newblock Rethinking imagenet pre-training.
\newblock In {\em ICCV}, 2019.

\bibitem{raghu2019transfusion}
Maithra Raghu, Chiyuan Zhang, Jon Kleinberg, and Samy Bengio.
\newblock Transfusion: Understanding transfer learning with applications to
  medical imaging.
\newblock In {\em ICCV}, 2019.

\bibitem{chen2019closer}
Wei-Yu Chen, Yen-Cheng Liu, Zsolt Kira, Yu-Chiang~Frank Wang, and Jia-Bin
  Huang.
\newblock A closer look at few-shot classification.
\newblock 2019.

\bibitem{chen2018lstd}
Hao Chen, Yali Wang, Guoyou Wang, and Yu~Qiao.
\newblock Lstd: A low-shot transfer detector for object detection.
\newblock In {\em AAAI}, 2018.

\bibitem{hamaguchi2019rare}
Ryuhei Hamaguchi, Ken Sakurada, and Ryosuke Nakamura.
\newblock Rare event detection using disentangled representation learning.
\newblock In {\em Proceedings of Conference on Computer Vision and Pattern
  Recognition}, 2019.

\bibitem{hendrycks2019using}
Dan Hendrycks, Kimin Lee, and Mantas Mazeika.
\newblock Using pre-training can improve model robustness and uncertainty.
\newblock In {\em ICML}, 2019.

\bibitem{bella2013effect}
Antonio Bella, C{\`e}sar Ferri, Jos{\'e} Hern{\'a}ndez-Orallo, and
  Mar{\'\i}a~Jos{\'e} Ram{\'\i}rez-Quintana.
\newblock On the effect of calibration in classifier combination.
\newblock {\em Applied intelligence}, 38(4):566--585, 2013.

\bibitem{lakshminarayanan2017simple}
Balaji Lakshminarayanan, Alexander Pritzel, and Charles Blundell.
\newblock Simple and scalable predictive uncertainty estimation using deep
  ensembles.
\newblock In {\em NIPS}, pages 6402--6413, 2017.

\bibitem{hendrycks2019deep}
Dan Hendrycks, Mantas Mazeika, and Thomas~G Dietterich.
\newblock Deep anomaly detection with outlier exposure.
\newblock In {\em ICLR}, 2019.

\bibitem{simonyan2015very}
Karen Simonyan and Andrew Zisserman.
\newblock Very deep convolutional networks for large-scale image recognition.
\newblock In {\em ICLR}, 2015.

\bibitem{bouchacourt2018multi}
Diane Bouchacourt, Ryota Tomioka, and Sebastian Nowozin.
\newblock Multi-level variational autoencoder: Learning disentangled
  representations from grouped observations.
\newblock In {\em AAAI}, 2018.

\bibitem{mathieu2016disentangling}
Michael~F Mathieu, Junbo~Jake Zhao, Junbo Zhao, Aditya Ramesh, Pablo
  Sprechmann, and Yann LeCun.
\newblock Disentangling factors of variation in deep representation using
  adversarial training.
\newblock In {\em NIPS}, 2016.

\bibitem{chan1998toward}
Philip~K Chan and Salvatore~J Stolfo.
\newblock Toward scalable learning with non-uniform class and cost
  distributions: A case study in credit card fraud detection.
\newblock In {\em KDD}, 1998.

\bibitem{yan2003predicting}
Rong Yan, Yan Liu, Rong Jin, and Alex Hauptmann.
\newblock On predicting rare classes with svm ensembles in scene
  classification.
\newblock In {\em ICASSP}, volume~3, pages III--21. IEEE, 2003.

\bibitem{he2016deep}
Kaiming He, Xiangyu Zhang, Shaoqing Ren, and Jian Sun.
\newblock Deep residual learning for image recognition.
\newblock In {\em CVPR}, pages 770--778, 2016.

\bibitem{lecun1998gradient}
Yann LeCun, L{\'e}on Bottou, Yoshua Bengio, Patrick Haffner, et~al.
\newblock Gradient-based learning applied to document recognition.
\newblock {\em IEEE}, 86(11):2278--2324, 1998.

\bibitem{kingma2015adam}
Diederik~P Kingma and Jimmy Ba.
\newblock Adam: A method for stochastic optimization.
\newblock In {\em ICLR}, 2015.

\bibitem{muller2019does}
Rafael M{\"u}ller, Simon Kornblith, and Geoffrey Hinton.
\newblock When does label smoothing help?
\newblock In {\em NeurIPS}, 2019.

\end{thebibliography}

\appendix

\newpage

\section{Samples from ABCD Dataset}
\label{app:abcd}

\begin{figure}[h]
\begin{adjustbox}{center}
\hfill
\includegraphics[width=\linewidth]{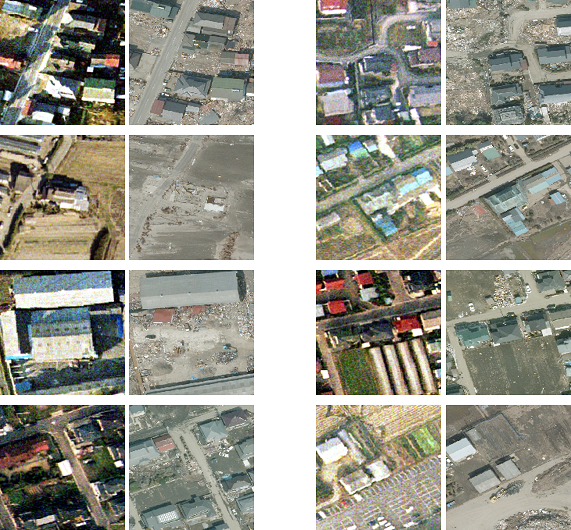}
\end{adjustbox}
\caption{Positive (Left) and negative (Right) samples from ABCD dataset.}
\label{fig:abcd}
\end{figure}

\newpage

\section{Codes for the Modified LeNet}
\label{app:arch}

\begin{lstfloat}
\begin{lstlisting}[language=Python, caption=Implementation of the Modified LeNet Architecture in PyTorch, label={code:imp}][t]
class LeNet(nn.Module):
    def __init__(self):
        super(LeNet, self).__init__()
        self.conv1 = nn.Conv2d(3, 6, kernel_size=5, stride=2)
        self.conv2 = nn.Conv2d(6, 16, kernel_size=5, stride=2)
        self.fc1 = nn.Linear(784, 120)
        self.fc2 = nn.Linear(120, 84)

    def forward(self, x):
        x = F.relu(self.conv1(x))
        x = F.max_pool2d(x, 2)
        x = F.relu(self.conv2(x))
        x = F.max_pool2d(x, 2)
        x = x.view(x.size(0), -1)
        x = F.relu(self.fc1(x))
        x = F.relu(self.fc2(x))
        return x


class Lenet_classifier(nn.Module):
    def __init__(self, feature_extractor):
        super(Lenet_classifier, self).__init__()
        self.feature_extractor = feature_extractor
        self.classifier_1 = nn.Linear(2 * 84, 128)
        self.classifier_2 = nn.Linear(128, 1)

    def forward(self, image_a, image_b):
        feature_a = self.feature_extractor(image_a)
        feature_b = self.feature_extractor(image_b)
        features = torch.cat((feature_a, feature_b), 1)

        out = self.classifier_1(features)
        out = F.relu(out)
        out = self.classifier_2(out)
        return out
\end{lstlisting}
\end{lstfloat}

\end{document}